\documentclass[sigconf]{acmart}

\renewcommand\footnotetextcopyrightpermission[1]{} 
\usepackage{enumerate}
\usepackage{multirow}
\usepackage{tabularx}
\usepackage{subfigure}
\def\BibTeX{{\rm B\kern-.05em{\sc i\kern-.025em b}\kern-.08emT\kern-.1667em\lower.7ex\hbox{E}\kern-.125emX}}
    
\setcopyright{none}

\begin{document}

%
\title{A High-Efficiency Framework for Constructing \\Large-Scale Face Parsing Benchmark}
%

\author{Yinglu Liu, Hailin Shi, Yue Si, Hao Shen, Xiaobo Wang and Tao Mei}
\affiliation{JD AI}

%

%
\begin{abstract}
Face parsing, which is to assign a semantic label to each pixel in face images, has recently attracted increasing interest due to its huge application potentials. 
Although many face related fields (\textit{e.g.} face recognition and face detection) have been well studied for many years, the existing datasets for face parsing are still severely limited in terms of the scale and quality, \textit{e.g.} the widely used Helen dataset only contains 2,330 images.
This is mainly because pixel-level annotation is a high cost and time-consuming work, especially for the facial parts without clear boundaries. The lack of accurate annotated datasets becomes a major obstacle in the progress of face parsing task.
It is a feasible way to utilize dense facial landmarks to guide the parsing annotation. However, annotating dense landmarks on human face encounters the same issues as the parsing annotation.
To overcome the above problems, in this paper, we develop a high-efficiency framework for face parsing annotation, which considerably simplifies and speeds up the parsing annotation by two consecutive modules. Benefit from the proposed framework, we construct a new Dense Landmark Guided Face Parsing (LaPa) benchmark~\footnote{http://hailin-ai.xyz/.}. It consists of 22,000 face images with large variations in expression, pose, occlusion, \textit{etc}. Each image is provided with accurate annotation of a 11-category pixel-level label map along with coordinates of 106-point landmarks. To the best of our knowledge, it is currently the largest public dataset for face parsing (almost ten times larger than other public datasets).
To make full use of our LaPa dataset with abundant face shape and boundary priors, we propose a simple yet effective Boundary-Sensitive Parsing Network (BSPNet). Our network is taken as a baseline model on the proposed LaPa dataset, and meanwhile, it achieves the state-of-the-art performance on the Helen dataset without resorting to extra face alignment.
We wish our work could help the community for further development on face parsing.
\end{abstract}

%
\keywords{face parsing, facial landmark localization, deep learning, dataset}

%
\begin{teaserfigure}
	\includegraphics[width=1\linewidth]{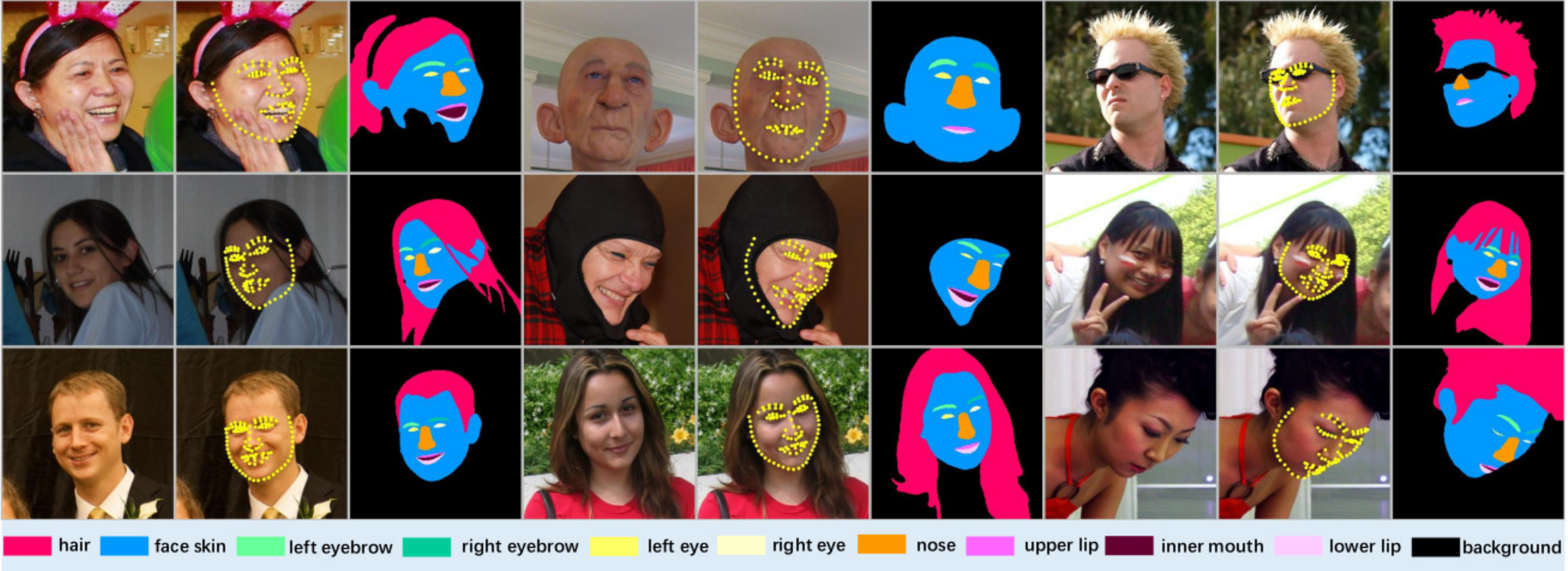}
	\caption{Annotation examples of the proposed LaPa dataset. It consists of 22,000 images with large variations in pose, facial expression, occlusion, \textit{etc}. Each image is annotated with the proposed semi-automatic labeling framework efficiently. It has obvious advantages on both annotation quality and efficiency compared with the existing face parsing collection. To the best of our knowledge, it is currently the largest public dataset for face parsing.} \label{fig:sample}
\end{teaserfigure}

%
\maketitle

\section{Introduction}
Face parsing, aiming to assign pixel-level semantic labels for face images, has attracted much attention due to its wide application potentials, such as facial beautification~\cite{ou2016beauty}, face image synthesis~\cite{zhang2018synthesis}, \textit{etc}. In recent years, deep learning promotes the development of artificial intelligence in computer vision and multimedia, especially on the face related fields. As is well known, adequate training data is crucial for achieving good results of deep learning methods. However, there are rare public datasets for face parsing due to the difficulty and high cost of pixel-level annotation. 

It is a feasible way to utilize the dense facial landmarks to guide the face parsing annotation. However, annotating tens of hundreds of landmarks encounters the same issues of the parsing annotation whereas well-trained annotators are needed. Therefore, the public datasets \cite{sagonas2013300,le2012interactive,ramanan2012face,gross2010multi,messer1999xm2vtsdb,wu2018look,koestinger2011annotated} for facial landmark localization are limited either on the number of training samples or the number of landmarks. Specifically, most of existing landmark datasets are annotated with less than 100 points, which are not enough to depict the shape of facial parts with fine details. For example, the widely used 68-point landmarks in 300W \cite{sagonas2013300} describe the eyebrow with only 5 points on the upper boundary while leaving the lower boundary unmarked. The recent 98-point landmarks in WFLW \cite{wu2018look} do not include the position of nose wing. Needless to say other makeups such as 21-point in AFLW \cite{koestinger2011annotated} and 6-point in AFW~\cite{baltrusaitis2013constrained}, they could be applied for face geometric normalization but are incompetent to represent boundaries of facial parts. In contrast, the Helen \cite{le2012interactive} dataset contains 194-point landmarks, but the number of samples is only 2,330 and no landmarks are located on the nose bridge.  

To remedy the above problems, in this paper we develop a high-efficiency framework for face parsing annotation, which is composed of two consecutive modules. 
In the first module, we develop a semi-automatic labeling tool for 106-point facial landmarks. With the help of an auxiliary landmark localization model, the coarse position for each landmark will be given firstly, so that annotators only need to adjust a small number of points in difficult cases as shown in Fig. \ref{fig:lm_labelingtool}. The initial landmarks play a role of strong and confident reference, and thus the annotators do not need heavy training to learn each landmark's definition. Benefit from these advantages, this tool significantly reduces the workload and speeds up the process of annotation for dense landmarks. 
In the second module, we propose a category-wise fitting approach, which draws the accurate contour for each facial part according to the landmarks from the first module; moreover, a coarse-to-fine segmentation strategy is employed to label the hair and face skin region. 
Finally, The outputs are merged hierarchically to produce a 11-category pixel-level semantic label map.
By using the proposed method, we construct a new benchmark for face parsing, named LaPa. It consists of 22,000 images, which cover large variations in facial expression, pose, occlusion, \textit{etc}. Each image is provided with accurate annotation of a 11-category pixel-level label map, namely hair, face skin, left/right eyebrow, left/right eye, nose, upper/lower lips, inner mouth and background along with the coordinates of 106-point landmarks. 

Furthermore, to make full use of our LaPa dataset with abundant face shape and boundary priors, we propose a simple yet effective Boundary-Sensitive Parsing Network (BSPNet), which improves the face parsing performance by focusing more on the boundary pixels from two aspects: 1) the boundary-aware features are integrated into semantic-aware features to preserve more boundary details implicitly. 2) the semantic loss of boundary pixels is weighted by the boundary map to reinforce the boundary effect explicitly. 
The experiments on the Helen and LaPa datasets demonstrate the effectiveness of our network.


The contribution of this paper are summarized as follows:
\begin{itemize}
    \item [1)]
    We develop a high-efficiency framework for face parsing annotation, which is composed of a Dense Landmark Annotation (DLA) module and a Pixel-Level Parsing Annotation (PPA) module. This framework considerably simplifies and speeds up the pixel-level parsing annotation.
    \item [2)]
    Based on the proposed framework, we construct a new large benchmark for face parsing. It contains 22,000 images. Each image is provided with a 11-category pixel-level label map 
    and coordinates of 106-point landmarks. To the best of our knowledge, this is the largest public dataset for face parsing so far. It will be released to the community soon.
    \item [3)]
    We propose an effective boundary-sensitive parsing network, which is taken as the baseline method on the proposed LaPa dataset. Meanwhile, we evaluate it on the public Helen dataset, and our model achieves the state-of-the-art performances on all categories.
\end{itemize}

\section{Related Work}
\subsection{Face Parsing}

\subsubsection{Dataset} Due to the aforementioned problems in pixel-level annotation, there are few face parsing datasets published. The most commonly used public datasets for face parsing methods are LFW-PL~\cite{kae2013augmenting} and Helen~\cite{le2012interactive,smith2013exemplar}.
LFW-PL is a subset of the Labeled Faces in the Wild (LFW) funneled images which is a database of face photographs dedicated to the unconstrained face recognition. This dataset contains 2,927 face images. All the images are first segmented into superpixels, and then each superpixel is manually assigned with one of the hair/skin/background categories. The annotations for facial parts are not provided in this dataset.
The original Helen dataset~\cite{le2012interactive} is composed of 2,330 face images with densely-sampled, manually-annotated keypoints around the semantic facial parts. Smith \textit{et al}. \cite{smith2013exemplar} generated segmentation ground truths of eye, eyebrow, nose, inside mouth, upper lip and lower lip automatically by using the contours, together with facial skin and hair categories generated from manually annotated boundaries and automatic matting algorithm~\cite{levin2008spectral}. 

\subsubsection{Methods} In recent years, increasing attention has been drawn in face parsing due to its great application potentials. Early works mainly focus on hand-crafted features and probabilistic graphical models. Warrell~\textit{et al}. \cite{warrell2009labelfaces} proposed to use priors to model facial structure and get facial parts labels through a Conditional Random Field (CRF). Smith \textit{et al}. \cite{smith2013exemplar} adopted SIFT features to select examplers and computed segmentation map of a test image by propagating labels from the aligned exemplar images. Kae \textit{et al}.~\cite{kae2013augmenting} combined CRF with a Restricted Boltzmann Machine (RBM) to model both local and global structures for face labeling. More recently, certain works attempt to tackle the face parsing task with the help of deep learning to break the performance bottleneck of traditional methods. Luo \textit{et al}.~\cite{luo2012hierarchical} proposed a hierarchical face parsing framework with Deep Belief Networks (DBNs) as facial parts and components detectors. Liu \textit{et al}.~\cite{liu2015multi} exploited a Convolution Neural Network (CNN) to model both unary likelihoods and pairwise label dependencies. Yamashita \textit{et al}. \cite{yamashita2015cost} proposed a weighted cost function to improve performances for certain classes like eyes. Jackson \textit{et al}.\cite{jackson2016cnn} proposed a two-stage parsing framework with Fully Convolutional Networks (FCNs). Liu \textit{et al}.~\cite{liu2017face} designed a light-weight network which combines a shallow CNN with a spatially variant Recurrent Neural Network (RNN) and a coarse-to-fine approach for accurate face parsing. Wei \textit{et al}.~\cite{wei2017learning} introduced an automatic method for selecting receptive fields and achieved accurate parsing results for face images. Guo \textit{et al}.~\cite{guo2018residual} adopted a prior mechanism to refine the Residual Encoder Decoder Netwrk (RED-Net), and achieved state-of-the-art performance on both LFW-LP and Helen datasets. 

\subsection{Facial Landmark Localization}
\subsubsection{Datasets} There are more public datasets for facial landmark localizatioon than that for face parsing. However, the existing datasets have limitations on either the scale of training samples or the number of landmarks. 
For example, the Helen dataset~\cite{zhou2013extensive} contains densely defined 194-point landmarks, whereas only 2,330 images are included. The AFLW dataset~\cite{koestinger2011annotated} contains about 25k annotated faces in real-world images while only annotating at most 21 landmarks for each image. Moreover, certain of them are captured under controlled conditions. For example, images in XM2VTS dataset~\cite{hasan2013localizing} are captured under laboratory conditions with the same illumination conditions and neutral expression. In order to remedy these shortcomings, IBUG~\footnote{https://ibug.doc.ic.ac.uk/home} built a new dataset called 300W \cite{sagonas2013300}, which consists of several datasets (AFW~\cite{baltrusaitis2013constrained}, Helen~\cite{zhou2013extensive}, LFPW~\cite{jaiswal2013guided}, IBUG~\cite{sagonas2013300}, \textit{etc}) and re-annotated them with 68-point landmarks. Wu \textit{et al}. \cite{wu2018look} released a new dataset called WFLW, which contains 10,000 faces and annotates 98 landmarks per face image.

\subsubsection{Methods} The methods for facial landmark localization mainly fall into two categories: Model based methods and Regression based methods, both of which push the state-of-the-art performance. 1) Model based methods usually build the basic model first, and then learn the transformation for each specific sample. For example, Cootes \textit{et al}. \cite{cootes1995active} proposed the Active Shape Model (ASM) approach. The first step in this approach is to train a mean shape model, represented by the concatenation of a sequence of landmarks, and then is to search the locations according the basic shape model and the local features for each landmark. In another work \cite{cootes2001active}, the Active Appearance Model (AAM) is proposed by adding a texture model besides the shape model. Besides, Zhu \textit{et al}. \cite{zhu2016face} proposed to fit a dense 3D Morphable Model (3DMM \cite{blanz2003face}) to the image via cascaded convolutional neural netowrks, which could be used to synthesize face images in profile views to provide abundant samples for training. 2) There are usually two strategies for regression based methods. One strategy is to directly regress the coordinates of landmarks from the input image. For example, Xiong \textit{et al}.~\cite{xiong2013supervised} proposed a Supervised Descent Method (SDM) which concatenates the SIFT features around each landmark as the shape-index feature and learn the regression matrix by minimizing a Non-linear Leaset Squares (NLS) function. Sun \textit{et al}. \cite{sun2013deep} firstly proposed a three-level cascaded CNN method to solve the facial landmark detection problem. In \cite{zhou2013extensive}, the 68 landmarks are divided into two categories (inner points and contour landmarks) and localized from coarse to fine. Yang \textit{et al}. \cite{zhang2014facial} adopted a multi-task method to learn the coordinates of the landmarks along with the attributes. Lai \textit{et al}. \cite{lai2018deep} proposed an end-to-end recurrent convolutional system for face alignment from coarse to fine. The other strategy is to generate heatmaps for each landmark respectively and the coordinates of landmarks could be obtained by a post-processing \cite{yang2017stacked,merget2018robust,bulat2017far,dong2018style}. For example, Yang \textit{et al}. \cite{yang2017stacked} proposed to adopt the hourglass network for facial landmark localization and achieved the first place in 300W challenge.
Heatmap regression methods usually achieve higher performance than coordinate regression methods, while the computational cost is usually higher than the latter one.

\section{High-efficiency Framework for Face Parsing Annotation}
\begin{figure}[htbp]
	\centering
	\includegraphics[width=0.9\linewidth]{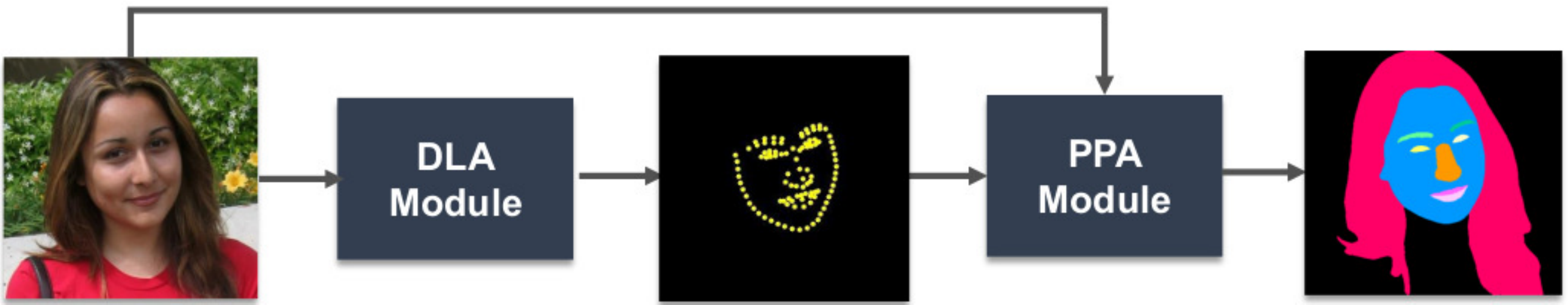}
	\caption{The proposed framework for face parsing. DLA denotes Dense Landmark Annotation and PPA denotes Pixel-level Parsing Annotation. First, DLA module outputs 106-point landmarks on the input color image, and then PPA module produces the pixel-level label map automatically according to the landmarks.} \label{fig:labeltool-fw}
\end{figure}

In this section, we will describe the proposed efficient framework for face parsing. As Fig.\ref{fig:labeltool-fw} shows, it is composed of two consecutive modules, named Dense Landmark Annotation (DLA) and Pixel-level Parsing Annotation (PPA). These two modules are introduced in Section \ref{DLA} and Section \ref{PPA} in details, respectively. In Section \ref{LaPa}, we introduce the information about the proposed LaPa dataset.

\subsection{Dense Landmark Annotation (DLA) Module}\label{DLA}
\begin{figure}[t]
	\centering
	\includegraphics[width=0.85\linewidth]{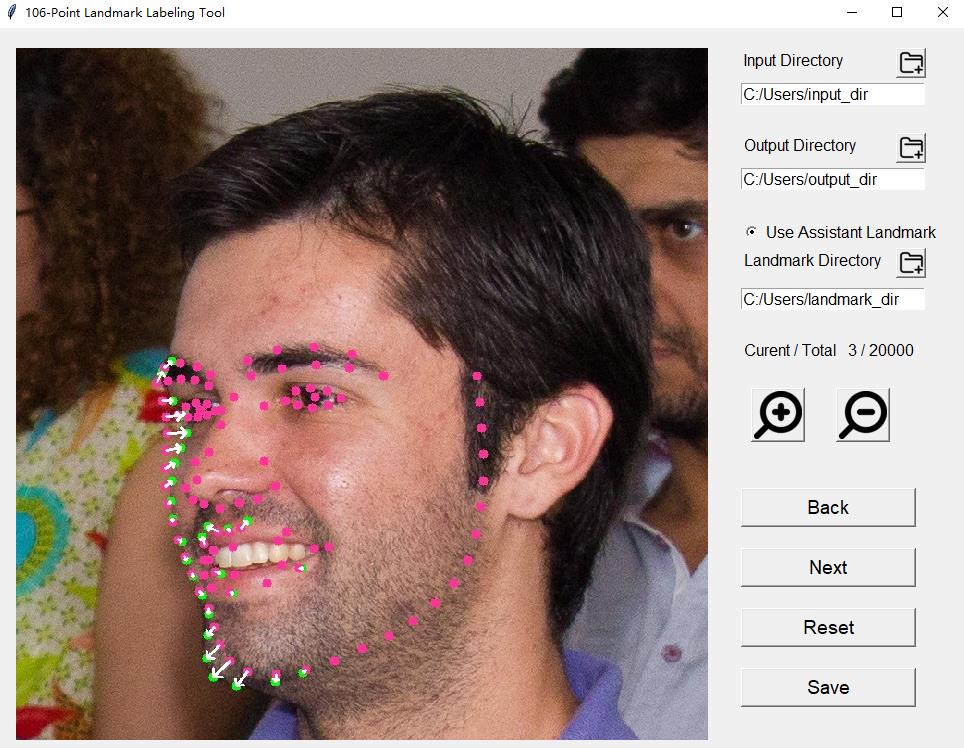}
	\caption{Our semi-automatic facial landmark labeling tool. The pink points refer to the initial results by the auxiliary model. The green points refer to the results after light manual correction. The white arrows show the adjustment trajectories. Our tool provides common functions such as undo, next, save, \textit{etc}. Best viewed in color and zoom.} \label{fig:lm_labelingtool}
\end{figure}

\begin{figure}[t]
    \centering
    \includegraphics[width=1\linewidth]{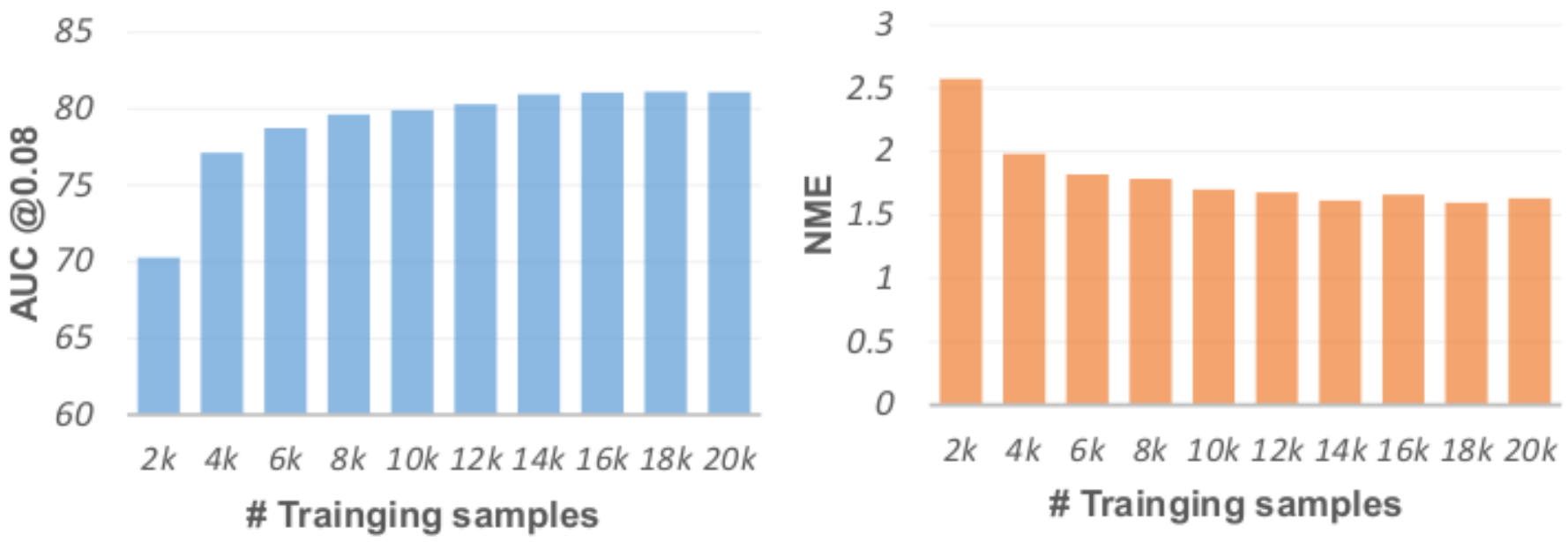}\label{fig:nme}
    \caption{Performance of the auxiliary model for landmark localization \textit{w.r.t.} the number of training samples. The horizontal axis refers to the number of training samples which is accumulated by our semi-auto facial landmark labeling tool. The vertical axis refers to the corresponding evaluation performance.}
    \label{fig:performance_hg}
\end{figure}

The purpose of DLA module is to annotate face images with dense landmarks efficiently. 
First, we develop a semi-automatic facial landmark labeling tool with user interface (Fig.\ref{fig:lm_labelingtool}). This tool can give a reference position for each landmark by an auxiliary facial landmark localization model, so that annotators only need to adjust a small number of points for difficult cases rather than annotating all from scratch. In this paper, 1-stack hourglass network \cite{newell2016stacked,Bulat_2017_ICCV} is employed , which is trained with mere 2,000 manually annotated images at the beginning and updated once 2,000 additional images are labeled. 
The Normalized Mean Error (NME) and Area Under Curve (AUC) with the number of training samples are reported in Fig.~\ref{fig:performance_hg}. We can see that the performance of the auxiliary model keeps improving along with the increasing samples accumulated by the semi-automatic labeling process. 
This tool significantly simplifies and speeds up the process of dense landmark annotation. In this work, the DLA module takes the 106-point landmark definition. The outputs of this module will be fed to the PPA module.

\subsection{Pixel-Level Parsing Annotation (PPA) Module}\label{PPA}

\begin{figure*}[htbp]
	\centering
	\includegraphics[width=0.9\linewidth]{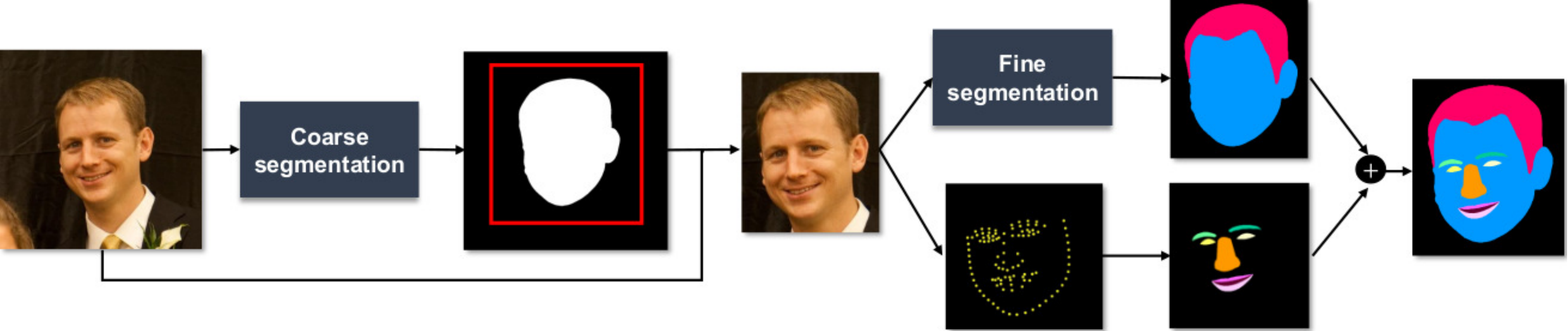}
	\caption{The framework for the proposed PPA module. First, the coarse segmentation is applied on the whole image and then the face region is cropped and finely segmented into three categories including hair, skin and background. Meanwhile, the annotation of facial parts are produced by our category-wise fitting approach according to the landmarks. Finally, the outputs are merged hierarchically as a complete annotation.} \label{fig:hp-flow}
\end{figure*}

This module takes a color image and coordinates of 106-point landmarks as input, and outputs a pixel-level semantic label map corresponding to 11 categories. As Fig. \ref{fig:hp-flow} shows, it consists of three stages:

1) \textbf{Coarse-to-fine segmentation for hair and face skin}

Parsing hair and skin are important for many facial applications, such as face beautification, hair coloring, \textit{etc}. However, conventional facial landmarks are not defined on the forehead and hair region, in part because forehead regions are usually covered by hair of different styles.
We solve this problem with the help of a human parsing dataset CIHP \footnote{http://sysu-hcp.net/lip/overview.php}. It is the first standard and comprehensive benchmark for instance-level human parsing. Because test set does not supply the ground truth, we just adopt the training and validation set, totally 33,280 images. 

In the coarse segmentation stage, we firstly map twenty categories in CIHP into two by taking hair and skin as foreground and others as background. Then we crop the regions of interest from original images according to the mapped labels incorporating with the instance labels. Usually, one image in CIHP could produce several sub-images in which only one major face exists. After filtering the images of which the width or height is less than 80 pixels, we collect about 26,000 images for training.
Here we adopt the advanced Pyramid Scene Parsing Network (PSPNet) \cite{zhao2017pyramid} to segment the foreground (hair and skin) from the background. 
This stage could be considered as a face detection operation to preserve the hair region while regular face detectors usually focus on the face region and may lose part of hair.

In the fine segmentation stage, we process the data in a similar way as the coarse segmentation stage but retaining hair and skin regions as two separate categories. In order to obtain more accurate segmentation results, the proposed Boundary-Sensitive Parsing Network (BSPnet) is adopted in this stage. The network will be introduced in details in Sec.~\ref{sec:BSPNet}.

2) \textbf{Category-wise fitting for facial parts}

\begin{figure}[htbp]
    \centering
    \includegraphics[width=0.7\linewidth]{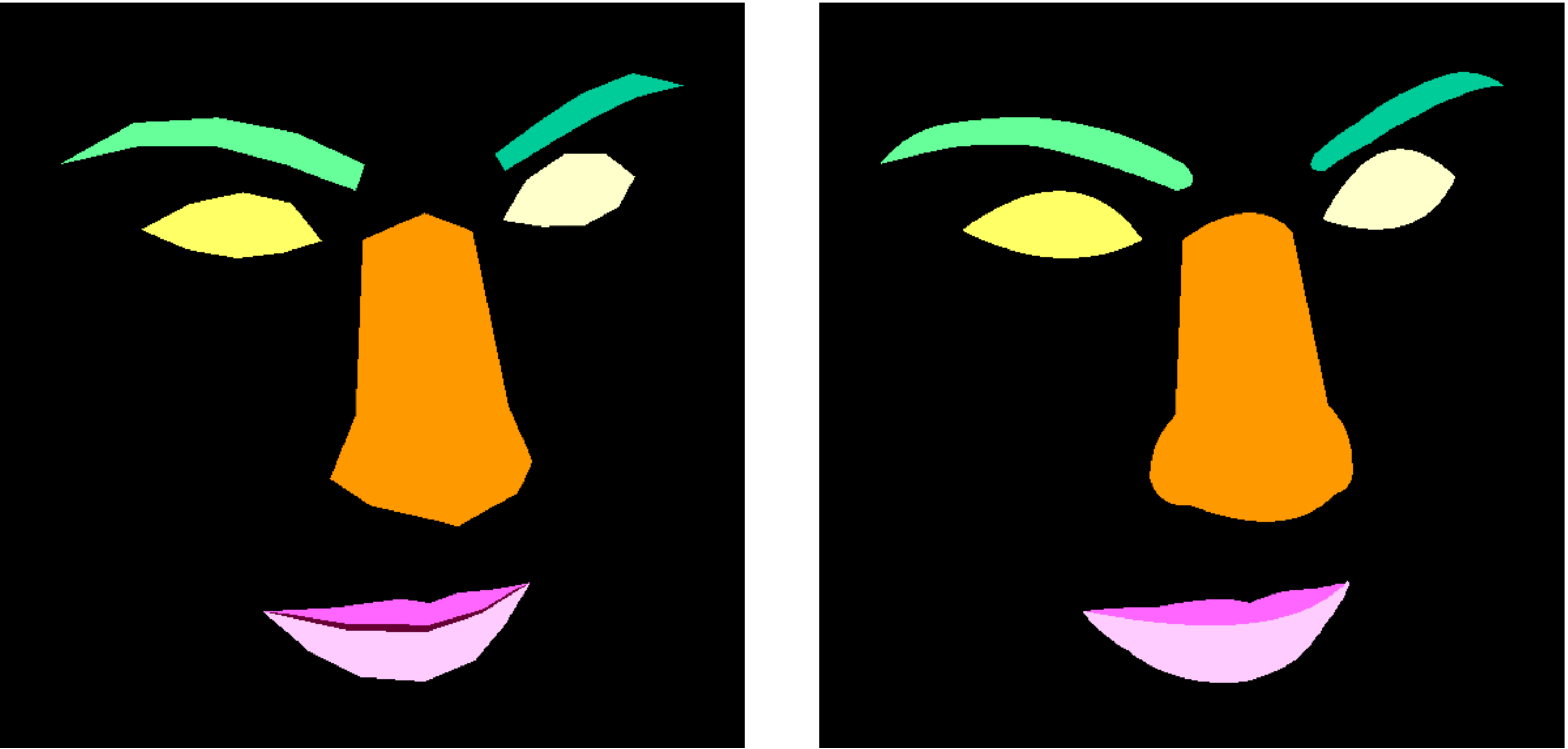}
    \caption{The effectiveness of category-wise fitting approach for facial parts. The left image shows the result by directly connecting neighboring landmarks in each category. The right image is the result by our method. Best viewed in zoom.}
    \label{fig:comp-fitting}
\end{figure}

Facial parts include left/right eyebrow, left/right eye, nose, upper/lower lip and inner mouth. In order to obtain more natural and accurate contours, we develop different fitting schemes for different facial parts according to their characteristics. For eyebrow, outer contour of mouth and jawline, we adopt polygon fitting to generate approximated contours. The pixels within each polygon are assigned to the corresponding category. In some cases, direct connection of long-distance neighboring landmarks may cause piecewise linear effect. To overcome this problem, prior knowledge is leveraged to make the results smoother by interpolation. For eye and inner mouth, two parabolas are applied to sketch the upper and lower boundary separately. For nose, we separate it into left and right parts to handle the profile case, and piecewise fitting is adopted due to the complex shape of nose. Note that all the partial landmarks are fitted in the transformed space where each part is aligned with a standard pose. The visualization results shown in Fig.~\ref{fig:comp-fitting} demonstrate the effectiveness of our approach.

3) \textbf{Fusion}

After the above two steps, we could obtain the label maps for hair/skin and facial parts. Then we merge them into a unified label map hierarchically. We emphasize that the order of fusion is important for producing correct results. For example, eye is always beyond the skin but sometimes behind hair or sunglasses, so if the label map of eye covers the label map of hair, the results will be unreasonable. Therefore, we merge the label maps in the order of skin, facial parts, hair and background.

\subsection{Dense Landmark Guided Face Parsing Benchmark}\label{LaPa}
We collect 22,000 images from two popular datasets-the landmark localization dataset 300W-LP \cite{sagonas2013300,zhu2016face} and the face recognition dataset Megaface \cite{kemelmacher2016megaface}. We randomly select 1000/2000 images from Megaface as validation/test set. The remaining images are taken as training set.
All the images are annotated by the DLA module first, and then the color image and the landmarks are fed to the PPA module to obtain 11-category pixel-level semantic annotation.
The label of each pixel denotes the semantic category according to the visible texture. Therefore, some categories may not present due to occlusion. For example, eye may be invisible due to large pose or occlusion by other objects such as sunglasses or hair. In this work, we just focus on single face parsing, and thus only the major face is annotated even if multiple faces exist in an image. Fig.\ref{fig:sample} shows some examples of the proposed LaPa dataset.

\begin{figure*}[t]
	\centering
	\includegraphics[width=0.93\linewidth]{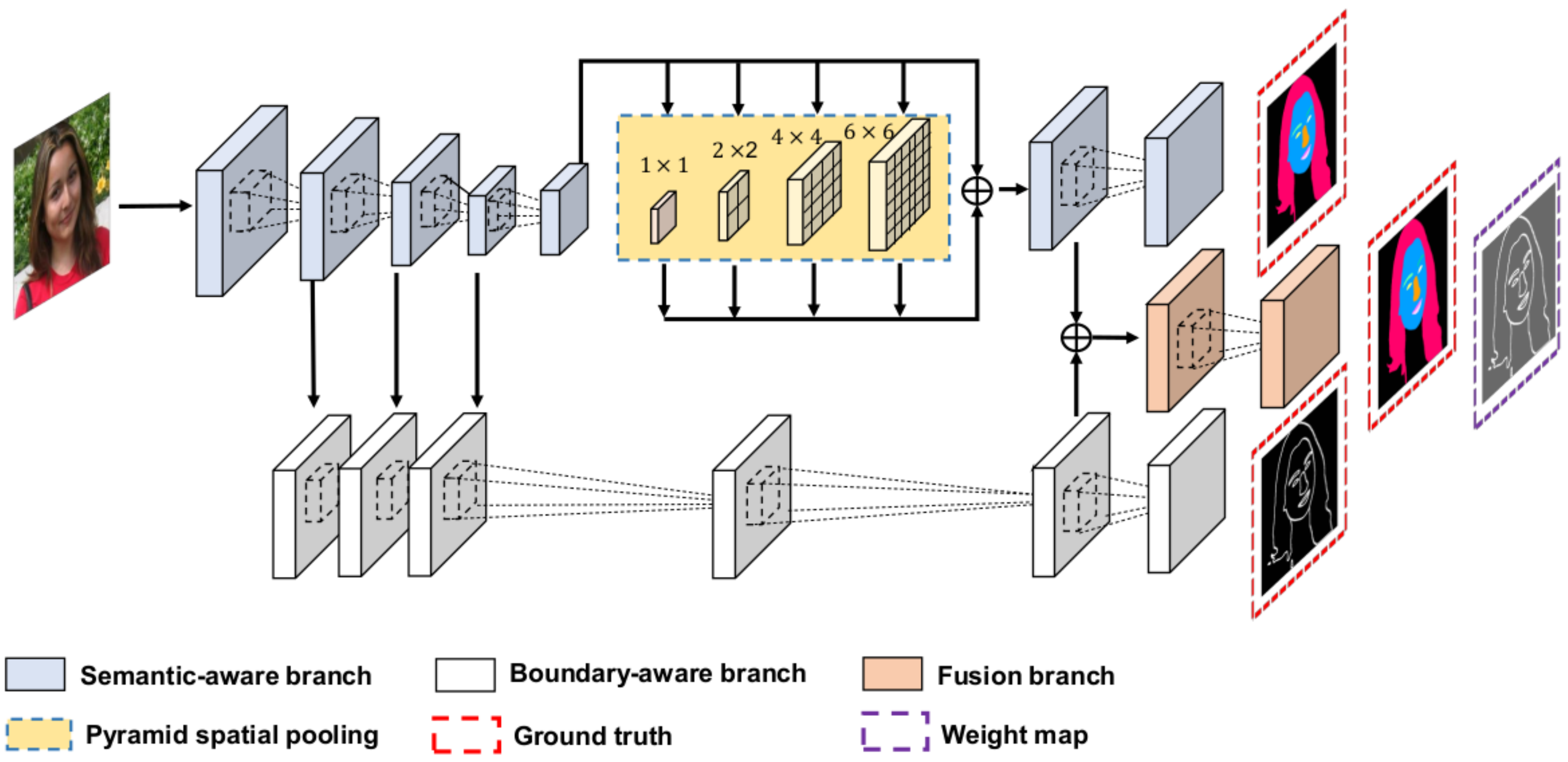}
	\caption{The network structure of BSPNet. It consists of three branches. The semantic-aware branch runs a multi-category semantic segmentation task. The boundary-aware branch runs a two-category boundary segmentation task. The fusion branch takes the combination of the features from the former two branches as input and employs the boundary map to weight the semantic segmentation loss.} \label{fig:BSPnet}
\end{figure*}

{\bfseries Comparison with relevant datasets.} \textbf{Helen}~\cite{smith2013exemplar} is a widely used dataset for face parsing, while it still has several limits: 1) The labeling is not accurate enough especially for hair and face skin categories produced by matting. As a result, most works based on Helen only focus on facial components, while ignoring hair and skin. 2) The limited number of samples and the lack of diversity in poses make it difficult to support training large-scale practical models.
The {\bfseries LFW-PL}~\cite{kae2013augmenting} dataset has the same issue of lack of training images while only hair and facial skin are annotated without considering facial parts. Compared to the existing datasets, the proposed LaPa dataset contains sufficient training samples which cover a wide range of variations, and provides annotation of fine-grained facial component categories, along with accurate hair and facial skin segments. Besides, with the advantage of the semi-automatic labeling framework, the LaPa dataset could be scaled up easily in future. Tab. \ref{tab:dataset-comp} gives statistics of the public datasets for face parsing. Meanwhile, we re-annotate the hair and facial skin categories on the Helen dataset by our framework introduced in Sec.~\ref{PPA}. The visual comparison results shown in Fig. \ref{fig:comp_on_helen} demonstrate the superiority of our framework.

\begin{table}[htbp]
	\centering 
	\caption{Statistics of the public datasets for face parsing. We show the number of images in training and test sets as well as the nubmer of categories including background.} 
	\label{tab:dataset-comp}
	\begin{tabular}{ccccc}  
		\hline
		Dataset&\#Training&\#Validation&\#Test&\#Category \\
		\hline
		LFW-PL~\cite{kae2013augmenting}&1500&500&927&3 \\
		\hline
		Helen~\cite{smith2013exemplar}&2000&230&100&11 \\
		\hline
		LaPa (Ours)&19000&1000&2000&11 \\
		\hline
	\end{tabular}
\end{table}

\section{Boundary-Sensitive Parsing Network} \label{sec:BSPNet}

Although face could be approximately considered as a rigid body in which the deformation is limited, face parsing is still difficult due to the variations in facial expression and pose. Furthermore, the parsing regions such as eye, nose, \textit{etc}. are usually smaller than general objects. All these reasons make it difficult to solve this specific task with general object segmentation or scene parsing methods \cite{zhao2017pyramid,long2015fully,badrinarayanan2017segnet,lin2017refinenet,chen2018deeplab,chen2017rethinking}.

To overcome the above limitations and make full use of our LaPa dataset, we propose a novel Boundary-Sensitive Parsing Network (BSPNet).
As Fig.\ref{fig:BSPnet} shows, the BSPNet consists of three branches.
The upper branch of the network is called semantic-aware branch. The purpose is to learn semantic-aware features and infer accurate semantic label maps from input images. Any existing segmentation network structure could be adopted in this branch. Here we employ ResNet-101~\cite{he2016deep} as the feature extraction backbone. In order to reduce the resolution loss caused by pooling or convolution with stride larger than 1, dilation convolution~\cite{chen2018deeplab} is adopted in the fifth residual block, therefore the resolution of the output is 1/16 rather than 1/32 of the input. In order to leverage the global texture information, pyramid spatial pooling with different scales are used before the classifier learning. Then, the output feature maps with different resolutions are concatenated with high-resolution feature maps generated by the last residual block after interpolation to the same scale. The integrated feature maps are used to predict the semantic label for each pixel.

Similar to \cite{CE2P2019}, the lower branch of the network is for boundary-aware feature learning, called boundary-aware branch. First, it extracts shared features from different layers of ResNet-101 in the semantic-aware branch, and projects them into a new space where boundary details are well preserved. The output of this branch is a boundary map in which each value refers to the confidence score that pixel is located on the boundary without considering semantics. The ground truths of this branch are computed according to the gradients of the label map. 

It is a common issue that the boundary pixels are difficult to distinguish due to confusion with the adjacent pixels belonging to different categories. 
Therefore, we further develop a fusion branch to boost the segmentation performance for these "hard samples". This branch takes the combination of features learned by the semantic- and boundary- aware branches as input, which are rich in semantics while boundary details are well preserved. As the same in semantic-aware branch, the output of fusion branch is a confidence map with $n$ channels, where $n$ denotes the number of semantic categories. Meanwhile, a weight map computed from the boundary map is used to enlarge the loss of boundary pixels.



The loss functions are defined as follows:

\begin{align}
    &L  = \lambda_1 L_s + \lambda_2 L_b + \lambda_3 L_{f}, \label{fun:loss} \\
    &L_s  = - \frac{1}{N} \sum_{i=1}^{N} \sum_{j=1}^{C} y^s_{ij} \log p^s_{ij}, \\
    &L_b  = - \frac{1}{N} \sum_{i=1}^{N} (y^b_i \log p^b_i + (1-y^b_i) \log (1-p^b_i)), \\
    &L_{f} = - \frac{1}{N} \sum_{i=1}^{N} \sum_{j=1}^{C} w_i\ y^s_{ij} \log p^{f}_{ij}, \label{fun:lf}
\end{align}
where $L$ refers to the total loss. $L_s$, $L_b$ and $L_f$ denote the loss of the semantic-aware, boundary-aware, and fusion branches, respectively. $\lambda_1$, $\lambda_2$ and $\lambda_3$ are hyper-parameters to balance the loss of different branches. $N$ denotes the number of pixels in the whole image while $C$ denotes the number of parsing categories. 
Here $y_{ij}^s$ equals to 1 if the semantic label of pixel $i$ is $j$, and $y_{ij}^s = 0$ otherwise. $y^b_i$ is a indicator variable of which the value is 1 if pixel $i$ is located on the boundaries and 0 otherwise. $p^s, p^b$ and $p^f$ are prediction values for the three branches, respectively. To enhance the effect of boundaries, we introduce a new parameter $w_i = 1 + \alpha$, if $y^b_i = 1$ and $w_i = 1$, otherwise. 
$\alpha$ is usually set to a positive number to increase the weight for boundary pixels.
During test phase, $p^{f}$ are taken as the output of the BSPNet.

\section{Experiments}
In this section, we first prove the effectiveness of the proposed BSPNet on the LaPa dataset. Then we evaluate our network on the public Helen dataset. Without utilizing any prior, our model achieves the best results over other state-of-the-art methods. 

\begin{table*}[htbp]
	\centering
	\caption{Ablation study on the LaPa dataset. \textbf{Model A} is trained only by the semantic-aware branch. \textbf{Model B} is trained by Model A plus the boundary-aware features. \textbf{Model C} is trained by Model B plus the boundary-aware weighted loss. The performances of each category, together with the mean F1-score over the 10 foreground categories are listed.} 
	\label{table1} 
	\begin{tabular}{ccccccccccccc}  
		\hline
		&\multirow{2}{*}{hair}&\multirow{2}{*}{skin}&left&right&left&right&\multirow{2}{*}{nose}&upper&inner&lower&\multirow{2}{*}{background}&\multirow{2}{*}{mean} \\ 
		&&&eyebrow&eyebrow&eye&eye&&lip&mouth&lip&& \\
		\hline
		\textbf{Model A}&94.86&95.95&81.79&81.61&81.50&81.69&93.79&80.10&84.10&80.45&98.76&85.58 \\
		\textbf{Model B}&95.30&96.27&83.32&82.82&82.45&82.72&94.43&81.25&84.73&81.24&98.86&86.45 \\
		\textbf{Model C} (\textbf{Baseline})&\textbf{95.32}&\textbf{96.54}&\textbf{84.34}&\textbf{84.27}&\textbf{84.86}&\textbf{85.17}&\textbf{94.66}&\textbf{82.45}&\textbf{85.63}&\textbf{82.31}&\textbf{98.86}&\textbf{87.55} \\
		\hline	
	\end{tabular}
\end{table*}
\begin{table*}[htbp]
	\centering  
	\caption{Comparison with state-of-the-art methods on the Helen dataset. To keep consistent with other methods, the performances of the hair category and other fine-grained categories (\textit{e.g.} left/right eyes) are not given. The overall scores are computed by combining the merged brows/eyes/mouth and nose categories. BSPNet+LaPa means the model pretrained on the LaPa dataset is employed as the model initialization.}
	\label{table3} 
	\begin{tabular}{cccccccccc}  
		\hline  
		&skin&nose&upper-lip&inner-mouth&lower-lip&brows&eyes&mouth&overall \\ 
		\hline 		
		Smith \textit{et al}. \cite{smith2013exemplar}&88.2&92.2&65.1&71.3&70.0&72.2&78.5&85.7&80.4\\
		Liu \textit{et al}. \cite{liu2015multi}&91.2&91.2&60.1&82.4&68.4&73.4&76.8&84.9&85.9 \\
		Liu \textit{et al}. \cite{liu2017face}&92.1&93.0&74.3&89.1&81.7&77.0&86.8&89.1&88.6\\
		Guo \textit{et al}. \cite{guo2018residual}&93.8&94.1&75.8&83.7&83.1&80.4&87.1&92.4&90.5\\
		\textbf{BSPNet}&\textbf{94.8}&\textbf{94.5}&\textbf{78.0}&\textbf{86.2}&\textbf{86.9}&\textbf{81.3}&\textbf{87.5}&\textbf{93.5}&\textbf{91.0}\\
		\textbf{BSPNet+LaPa}&\textbf{95.1}&\textbf{94.7}&\textbf{80.2}&\textbf{86.6}&\textbf{86.9}&\textbf{81.9}&\textbf{87.8}&\textbf{93.8}&\textbf{91.4}\\
		\hline
	\end{tabular}
\end{table*}

\subsection{Experimental Setting}
For both LaPa and Helen datasets, we adopt similar network configurations. 
For the semantic-aware branch, the network parameters of ResNet-101 are initialized from the model pretrained on the ImageNet dataset \cite{deng2009imagenet}. The input size of the network is 473$\times$473, and dilation convolution is used in the last residual block to retain the feature resolution. The extracted features are then processed by a spatial pyramid pooling module with four different scales of $1\times1$, $2\times2$, $3\times3$ and $6\times6$ to aggregate global and local contextual information. 
For the boundary-aware branch, the feature maps from conv2\_3, conv3\_4 and conv4\_23 in ResNet-101 are concatenated as input. As the same in \cite{CE2P2019}, we adopt a positive/negative sample balancing strategy which takes the ratio of pixels belonging to specific class as the weights of the opposite one.
For the fusion branch, the last feature maps before predictors of the semantic- and boundary- aware branches are concatenated as the input features. The ground truth of this branch is the same as semantic-aware branch, while the ground truth of the boundary-aware branch is adopted to generate the weight map.

The network is trained by minimizing the objective function defined in Eq. (\ref{fun:loss}). We use mini-batch gradient descent as the optimizer with the momentum of 0.9, weight decay of 0.0005 and batch size of 64. "Poly" learning rate policy is used to update parameters and the initial learning rate is set to 0.001. Synchronized Batch Normalization is adopted to accelerate the training procedure.
The $\lambda1$, $\lambda2$ and $\lambda3$ are set to 1, 1, and 2 respectively. $\alpha$ is set to 200. All the hyper-parameters are determined on the validation set. 
Our experiments are implemented by Pytorch framework, and all the models are trained on 4 NVIDIA Tesla P40 with 24GB memory.
Like \cite{smith2013exemplar,liu2015multi,guo2018residual}, We adopt F1-score as the quantitative evaluation metric, which is the harmonic mean of precision and recall. 

\begin{align}
    F1 = 2 \times \frac{precision \cdot recall}{precision + recall}
\end{align}

\begin{figure*}[htbp]
	\centering
	\includegraphics[width=0.91\linewidth]{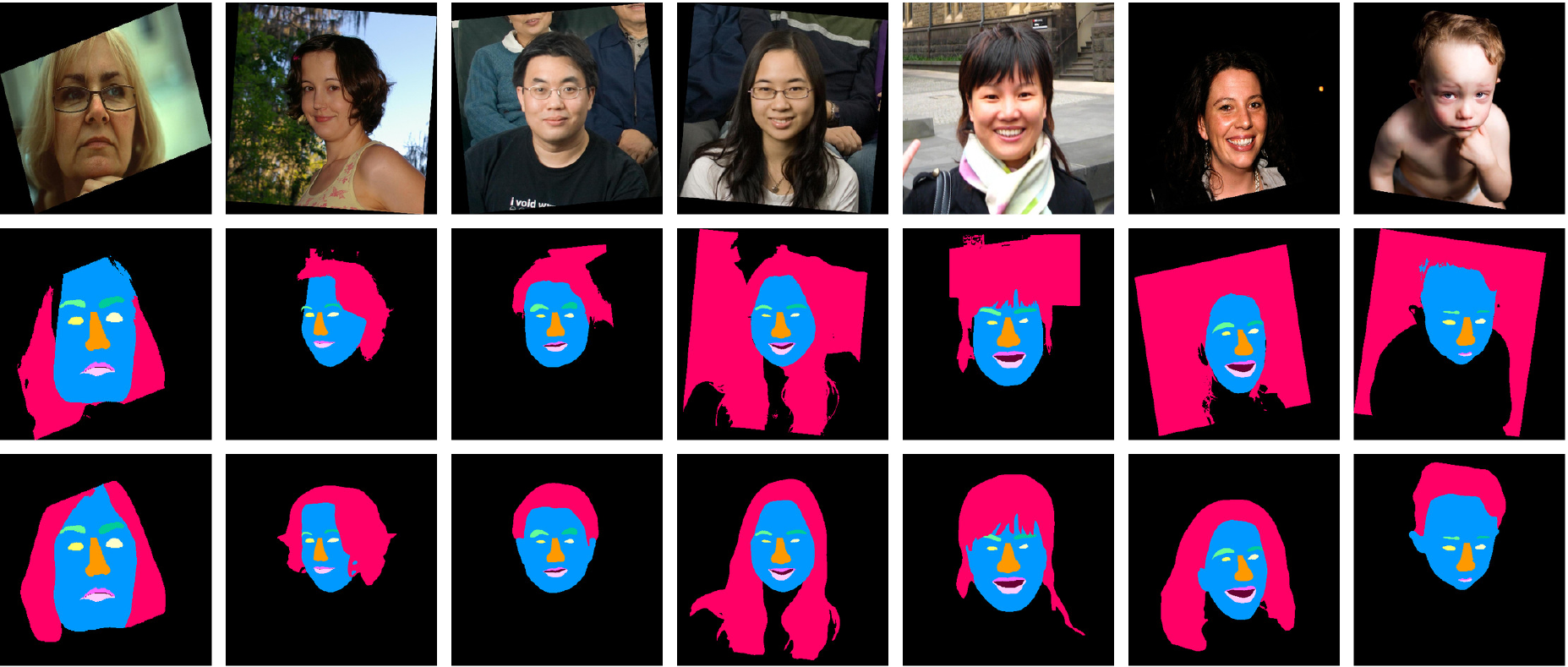}
	\caption{Comparison results on the Helen dataset. The first row refers to the original color image. The second row is the ground truth provided by Smith \textit{et al}. \cite{smith2013exemplar}. The third row shows the results where the hair and face skin categories are re-annotated by the proposed framework.} \label{fig:comp_on_helen}
\end{figure*}
\begin{figure}[htbp]
	\centering
	\includegraphics[width=1\linewidth]{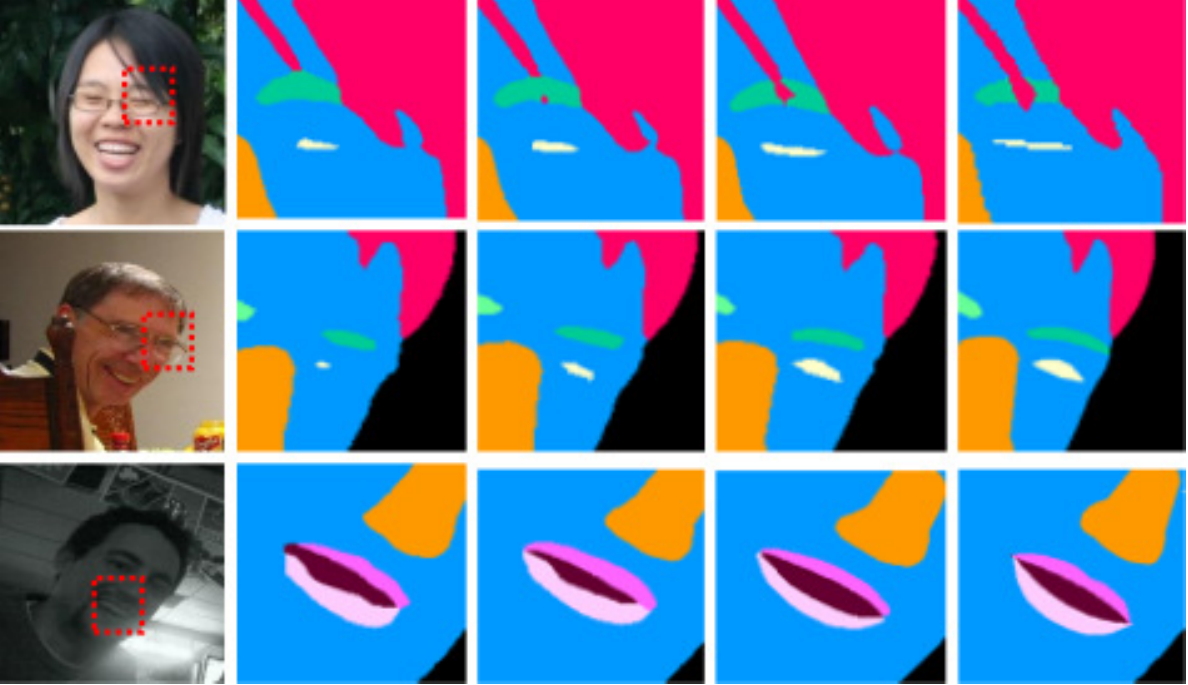}
	\caption{Visualizing the results on the LaPa dataset. The first column is the original image, in which the red dashed rectangles indicate the challenging parts. The next three columns are zoom-in results of the comparison methods (\textit{i.e.,} \textbf{Model A}, \textbf{Model B} and \textbf{Model C}). The last column shows the ground truth of the corresponding images.} \label{fig:ablation-comp}
\end{figure}

\subsection{Ablation Study} \label{sec:ablation}

To evaluate the effectiveness of the proposed network, three kinds of models are trained on the LaPa dataset. 
\textbf{Model A} is trained only by the semantic-aware branch, which does not utilize any auxiliary boundary information.
\textbf{Model B} is trained by all three branches without additional loss weights for boundary pixels, which is equivalent to set $w_i=1$ for all pixels in Eq. (\ref{fun:lf}).
\textbf{Model C} is trained by the proposed BSPNet, which utilizes both the boundary-aware features and boundary-aware weighted loss.
As Tab.~\ref{table1} shows, the performances of Model B are consistently better than that of Model A on all categories, while Model C achieves the best results over other models.
Specifically, the mean F1-score over 10 semantic categories (without background) of Model B is 86.45\%, improved by 0.87\% over Model A, while Model C further improves the accuracy to 87.55\%, 1.97\% and 1.1\% higher than Model A and Model B, respectively.
For hair, face skin and nose categories, of which the sizes are relatively large, Model C achieves 0.46\%, 0.59\% and 0.87\% improvements over Model A, respectively.
For the small-size categories of left/right eyebrow, left/right eye, upper/lower lip and inner mouth, the performances of Model C are significantly improved by 2.55\%/2.66\%, 3.36\%/3.48\%, 2.35\%/1.86\% and 1.53\% over Model A, respectively.
Fig.~\ref{fig:ablation-comp} shows the visualization results of the three models.
The experimental results demonstrate that the proposed BSPNet is effective for face parsing, especially for categories with small sizes.
The performance of our BSPNet (Model C in Tab.~\ref{table1}) could be taken as a baseline result for further researches on the proposed LaPa benchmark. 

\subsection{Comparison with State-of-the-art Methods}

\subsubsection{Dataset and ground truth}
Since the original Helen dataset~\cite{le2012interactive} is made for facial feature localization instead of parsing, Smith \textit{et al}.~\cite{smith2013exemplar} take several steps to convert densely labeled landmarks into segmentation maps. Specifically, ground truth segments of facial parts are automatically generated from manually-annotated contours. For facial skin, the jawline contour is used as the lower boundary, while for the upper boundary, an automatic matting algorithm~\cite{levin2008spectral} is used to separate the forehead from hair. The same matting strategy is adopted to recover the hair region. To make the dataset adaptive to semantic segmentation tasks, we first transform the confidence maps ranging from 0 to 255 to label maps by selecting the category with the maximum confidence value. As Fig. \ref{fig:comp_on_helen} shows, this will cause incorrect ground truth for some cases, especially for the hair category. 
As the number of training images are limited, many methods adopt exemplar based approach, which needs to split some images as exemplar during the training and test phase. As in \cite{liu2015multi,guo2018residual}, 230 images are splited as exemplars. In contrast, our method could directly output the prediction result for each pixel by the network, thus we use all the training images and exemplar images for training model, and test images are the same as \cite{liu2015multi,guo2018residual}.

\subsubsection{Experimental results}
As the previous works \cite{smith2013exemplar, liu2015multi,liu2017face, guo2018residual} do not report the performance of hair and fine-grained (\textit{i.e.} left/right eyebrow and left/right eye) categories on the Helen dataset, the mean score of foreground categories cannot be computed as Tab.~\ref{table1} shows.
To keep consistent with the previous methods, we report our results on the skin, nose, upper-lip, inner-mouth, lower-lip, merged brows, merged eyes and merged mouth categories. The overall scores are computed by combining the merged brows, merged eyes, merged mouth and nose categories without considering fine-grained categories.
As Tab.~\ref{table3} shows, our model achieves the best results over other state-of-the-art methods on all categories.
We emphasize that the performances of upper-lip, inner-mouth and lower-lip categories by ours are significantly improved by 2.2\%, 2.5\% and 3.8\% over Guo \textit{et al}.~\cite{guo2018residual}, while the accuracy of the merged mouth category is only 1.1\% higher than Guo \textit{et al}.~\cite{guo2018residual}.
That indicates the categories of upper lip, inner mouth and lower lip are severely confused by Guo \textit{et al.}~\cite{guo2018residual} while the merged accuracy ignores the confusion among them.
Our model achieves the best overall score of 91.0\%, outperforming Smith \textit{et al}. \cite{smith2013exemplar}, Liu \textit{et al}. \cite{liu2015multi}, Liu \textit{et al}. \cite{liu2017face} and Guo \textit{et al}.~\cite{guo2018residual} by 10.6\%, 5.1\%, 2.4\% and 0.5\%, respectively.
In addition, we employ the model pretrained on the LaPa dataset as the model initialization, and finetune it on the Helen dataset. We can see that the accuracy of each category is further improved, and the overall score is achieved by 91.4\%, which demonstrate the superiority of our dataset.

\section{Conclusion}
In this paper, we develop a high-efficiency framework for face parsing annotation, which significantly simplifies the pixel-level semantic annotation for face parsing with high accuracy. Benefit from this novel framework, we construct a new benchmark for face parsing. It consists of 22,000 face images and each image is provided with a 11-category semantic label map along with coordinates of 106-point landmarks. To the best of our knowledge, this is the largest public dataset for face parsing so far. Furthermore, we propose a simple yet effective boundary-sensitive parsing network, which boosts the segmentation performance by integrating boundary-aware features implicitly and weighting boundary-pixel loss explicitly. Experiments on Helen and the proposed LaPa datasets demonstrate the effectiveness of our network.

%
\bibliographystyle{ACM-Reference-Format}
\bibliography{ref}


\begin{thebibliography}{47}


\ifx \showCODEN    \undefined \def \showCODEN     #1{\unskip}     \fi
\ifx \showDOI      \undefined \def \showDOI       #1{#1}\fi
\ifx \showISBNx    \undefined \def \showISBNx     #1{\unskip}     \fi
\ifx \showISBNxiii \undefined \def \showISBNxiii  #1{\unskip}     \fi
\ifx \showISSN     \undefined \def \showISSN      #1{\unskip}     \fi
\ifx \showLCCN     \undefined \def \showLCCN      #1{\unskip}     \fi
\ifx \shownote     \undefined \def \shownote      #1{#1}          \fi
\ifx \showarticletitle \undefined \def \showarticletitle #1{#1}   \fi
\ifx \showURL      \undefined \def \showURL       {\relax}        \fi
\providecommand\bibfield[2]{#2}
\providecommand\bibinfo[2]{#2}
\providecommand\natexlab[1]{#1}
\providecommand\showeprint[2][]{arXiv:#2}

\bibitem[\protect\citeauthoryear{Badrinarayanan, Kendall, and
  Cipolla}{Badrinarayanan et~al\mbox{.}}{2017}]%
        {badrinarayanan2017segnet}
\bibfield{author}{\bibinfo{person}{Vijay Badrinarayanan}, \bibinfo{person}{Alex
  Kendall}, {and} \bibinfo{person}{Roberto Cipolla}.}
  \bibinfo{year}{2017}\natexlab{}.
\newblock \showarticletitle{Segnet: A deep convolutional encoder-decoder
  architecture for image segmentation}.
\newblock \bibinfo{journal}{\emph{IEEE transactions on pattern analysis and
  machine intelligence}} \bibinfo{volume}{39}, \bibinfo{number}{12}
  (\bibinfo{year}{2017}), \bibinfo{pages}{2481--2495}.
\newblock


\bibitem[\protect\citeauthoryear{Baltrusaitis, Robinson, and
  Morency}{Baltrusaitis et~al\mbox{.}}{2013}]%
        {baltrusaitis2013constrained}
\bibfield{author}{\bibinfo{person}{Tadas Baltrusaitis}, \bibinfo{person}{Peter
  Robinson}, {and} \bibinfo{person}{Louis-Philippe Morency}.}
  \bibinfo{year}{2013}\natexlab{}.
\newblock \showarticletitle{Constrained local neural fields for robust facial
  landmark detection in the wild}. In \bibinfo{booktitle}{\emph{Proceedings of
  the IEEE International Conference on Computer Vision Workshops}}.
  \bibinfo{pages}{354--361}.
\newblock


\bibitem[\protect\citeauthoryear{Blanz and Vetter}{Blanz and Vetter}{2003}]%
        {blanz2003face}
\bibfield{author}{\bibinfo{person}{Volker Blanz} {and} \bibinfo{person}{Thomas
  Vetter}.} \bibinfo{year}{2003}\natexlab{}.
\newblock \showarticletitle{Face recognition based on fitting a 3d morphable
  model}.
\newblock \bibinfo{journal}{\emph{IEEE Transactions on pattern analysis and
  machine intelligence}} \bibinfo{volume}{25}, \bibinfo{number}{9}
  (\bibinfo{year}{2003}), \bibinfo{pages}{1063--1074}.
\newblock


\bibitem[\protect\citeauthoryear{Bulat and Tzimiropoulos}{Bulat and
  Tzimiropoulos}{2017}]%
        {bulat2017far}
\bibfield{author}{\bibinfo{person}{Adrian Bulat} {and}
  \bibinfo{person}{Georgios Tzimiropoulos}.} \bibinfo{year}{2017}\natexlab{}.
\newblock \showarticletitle{How far are we from solving the 2d \& 3d face
  alignment problem?(and a dataset of 230,000 3d facial landmarks)}. In
  \bibinfo{booktitle}{\emph{Proceedings of the IEEE International Conference on
  Computer Vision}}. \bibinfo{pages}{1021--1030}.
\newblock


\bibitem[\protect\citeauthoryear{Chen, Papandreou, Kokkinos, Murphy, and
  Yuille}{Chen et~al\mbox{.}}{2018}]%
        {chen2018deeplab}
\bibfield{author}{\bibinfo{person}{Liang-Chieh Chen}, \bibinfo{person}{George
  Papandreou}, \bibinfo{person}{Iasonas Kokkinos}, \bibinfo{person}{Kevin
  Murphy}, {and} \bibinfo{person}{Alan~L Yuille}.}
  \bibinfo{year}{2018}\natexlab{}.
\newblock \showarticletitle{Deeplab: Semantic image segmentation with deep
  convolutional nets, atrous convolution, and fully connected crfs}.
\newblock \bibinfo{journal}{\emph{IEEE transactions on pattern analysis and
  machine intelligence}} \bibinfo{volume}{40}, \bibinfo{number}{4}
  (\bibinfo{year}{2018}), \bibinfo{pages}{834--848}.
\newblock


\bibitem[\protect\citeauthoryear{Chen, Papandreou, Schroff, and Adam}{Chen
  et~al\mbox{.}}{2017}]%
        {chen2017rethinking}
\bibfield{author}{\bibinfo{person}{Liang-Chieh Chen}, \bibinfo{person}{George
  Papandreou}, \bibinfo{person}{Florian Schroff}, {and}
  \bibinfo{person}{Hartwig Adam}.} \bibinfo{year}{2017}\natexlab{}.
\newblock \showarticletitle{Rethinking atrous convolution for semantic image
  segmentation}.
\newblock \bibinfo{journal}{\emph{arXiv preprint arXiv:1706.05587}}
  (\bibinfo{year}{2017}).
\newblock


\bibitem[\protect\citeauthoryear{Cootes, Edwards, and Taylor}{Cootes
  et~al\mbox{.}}{2001}]%
        {cootes2001active}
\bibfield{author}{\bibinfo{person}{Timothy~F Cootes}, \bibinfo{person}{Gareth~J
  Edwards}, {and} \bibinfo{person}{Christopher~J Taylor}.}
  \bibinfo{year}{2001}\natexlab{}.
\newblock \showarticletitle{Active appearance models}.
\newblock \bibinfo{journal}{\emph{IEEE Transactions on Pattern Analysis \&
  Machine Intelligence}} \bibinfo{number}{6} (\bibinfo{year}{2001}),
  \bibinfo{pages}{681--685}.
\newblock


\bibitem[\protect\citeauthoryear{Cootes, Taylor, Cooper, and Graham}{Cootes
  et~al\mbox{.}}{1995}]%
        {cootes1995active}
\bibfield{author}{\bibinfo{person}{Timothy~F Cootes},
  \bibinfo{person}{Christopher~J Taylor}, \bibinfo{person}{David~H Cooper},
  {and} \bibinfo{person}{Jim Graham}.} \bibinfo{year}{1995}\natexlab{}.
\newblock \showarticletitle{Active shape models-their training and
  application}.
\newblock \bibinfo{journal}{\emph{Computer vision and image understanding}}
  \bibinfo{volume}{61}, \bibinfo{number}{1} (\bibinfo{year}{1995}),
  \bibinfo{pages}{38--59}.
\newblock


\bibitem[\protect\citeauthoryear{Deng, Dong, Socher, Li, Li, and Fei-Fei}{Deng
  et~al\mbox{.}}{2009}]%
        {deng2009imagenet}
\bibfield{author}{\bibinfo{person}{Jia Deng}, \bibinfo{person}{Wei Dong},
  \bibinfo{person}{Richard Socher}, \bibinfo{person}{Li-Jia Li},
  \bibinfo{person}{Kai Li}, {and} \bibinfo{person}{Li Fei-Fei}.}
  \bibinfo{year}{2009}\natexlab{}.
\newblock \showarticletitle{Imagenet: A large-scale hierarchical image
  database}. In \bibinfo{booktitle}{\emph{2009 IEEE conference on computer
  vision and pattern recognition}}. Ieee, \bibinfo{pages}{248--255}.
\newblock


\bibitem[\protect\citeauthoryear{Dong, Yan, Ouyang, and Yang}{Dong
  et~al\mbox{.}}{2018}]%
        {dong2018style}
\bibfield{author}{\bibinfo{person}{Xuanyi Dong}, \bibinfo{person}{Yan Yan},
  \bibinfo{person}{Wanli Ouyang}, {and} \bibinfo{person}{Yi Yang}.}
  \bibinfo{year}{2018}\natexlab{}.
\newblock \showarticletitle{Style aggregated network for facial landmark
  detection}. In \bibinfo{booktitle}{\emph{Proceedings of the IEEE Conference
  on Computer Vision and Pattern Recognition}}. \bibinfo{pages}{379--388}.
\newblock


\bibitem[\protect\citeauthoryear{Gross, Matthews, Cohn, Kanade, and
  Baker}{Gross et~al\mbox{.}}{2010}]%
        {gross2010multi}
\bibfield{author}{\bibinfo{person}{Ralph Gross}, \bibinfo{person}{Iain
  Matthews}, \bibinfo{person}{Jeffrey Cohn}, \bibinfo{person}{Takeo Kanade},
  {and} \bibinfo{person}{Simon Baker}.} \bibinfo{year}{2010}\natexlab{}.
\newblock \showarticletitle{Multi-pie}.
\newblock \bibinfo{journal}{\emph{Image and Vision Computing}}
  \bibinfo{volume}{28}, \bibinfo{number}{5} (\bibinfo{year}{2010}),
  \bibinfo{pages}{807--813}.
\newblock


\bibitem[\protect\citeauthoryear{Guo, Kim, Zhang, Qian, Yoo, Xu, Zou, Han, and
  Choi}{Guo et~al\mbox{.}}{2018}]%
        {guo2018residual}
\bibfield{author}{\bibinfo{person}{Tianchu Guo}, \bibinfo{person}{Youngsung
  Kim}, \bibinfo{person}{Hui Zhang}, \bibinfo{person}{Deheng Qian},
  \bibinfo{person}{ByungIn Yoo}, \bibinfo{person}{Jingtao Xu},
  \bibinfo{person}{Dongqing Zou}, \bibinfo{person}{Jae-Joon Han}, {and}
  \bibinfo{person}{Changkyu Choi}.} \bibinfo{year}{2018}\natexlab{}.
\newblock \showarticletitle{Residual Encoder Decoder Network and Adaptive Prior
  for Face Parsing}. In \bibinfo{booktitle}{\emph{Thirty-Second AAAI Conference
  on Artificial Intelligence}}.
\newblock


\bibitem[\protect\citeauthoryear{Hasan, Pal, and Moalem}{Hasan
  et~al\mbox{.}}{2013}]%
        {hasan2013localizing}
\bibfield{author}{\bibinfo{person}{Md Hasan}, \bibinfo{person}{Christopher
  Pal}, {and} \bibinfo{person}{Sharon Moalem}.}
  \bibinfo{year}{2013}\natexlab{}.
\newblock \showarticletitle{Localizing facial keypoints with global descriptor
  search, neighbour alignment and locally linear models}. In
  \bibinfo{booktitle}{\emph{Proceedings of the IEEE International Conference on
  Computer Vision Workshops}}. \bibinfo{pages}{362--369}.
\newblock


\bibitem[\protect\citeauthoryear{He, Zhang, Ren, and Sun}{He
  et~al\mbox{.}}{2016}]%
        {he2016deep}
\bibfield{author}{\bibinfo{person}{Kaiming He}, \bibinfo{person}{Xiangyu
  Zhang}, \bibinfo{person}{Shaoqing Ren}, {and} \bibinfo{person}{Jian Sun}.}
  \bibinfo{year}{2016}\natexlab{}.
\newblock \showarticletitle{Deep residual learning for image recognition}. In
  \bibinfo{booktitle}{\emph{Proceedings of the IEEE conference on computer
  vision and pattern recognition}}. \bibinfo{pages}{770--778}.
\newblock


\bibitem[\protect\citeauthoryear{Jackson, Valstar, and Tzimiropoulos}{Jackson
  et~al\mbox{.}}{2016}]%
        {jackson2016cnn}
\bibfield{author}{\bibinfo{person}{Aaron~S Jackson}, \bibinfo{person}{Michel
  Valstar}, {and} \bibinfo{person}{Georgios Tzimiropoulos}.}
  \bibinfo{year}{2016}\natexlab{}.
\newblock \showarticletitle{A CNN cascade for landmark guided semantic part
  segmentation}. In \bibinfo{booktitle}{\emph{European Conference on Computer
  Vision}}. Springer, \bibinfo{pages}{143--155}.
\newblock


\bibitem[\protect\citeauthoryear{Jaiswal, Almaev, and Valstar}{Jaiswal
  et~al\mbox{.}}{2013}]%
        {jaiswal2013guided}
\bibfield{author}{\bibinfo{person}{Shashank Jaiswal}, \bibinfo{person}{Timur
  Almaev}, {and} \bibinfo{person}{Michel Valstar}.}
  \bibinfo{year}{2013}\natexlab{}.
\newblock \showarticletitle{Guided unsupervised learning of mode specific
  models for facial point detection in the wild}. In
  \bibinfo{booktitle}{\emph{Proceedings of the IEEE International Conference on
  Computer Vision Workshops}}. \bibinfo{pages}{370--377}.
\newblock


\bibitem[\protect\citeauthoryear{Kae, Sohn, Lee, and Learned-Miller}{Kae
  et~al\mbox{.}}{2013}]%
        {kae2013augmenting}
\bibfield{author}{\bibinfo{person}{Andrew Kae}, \bibinfo{person}{Kihyuk Sohn},
  \bibinfo{person}{Honglak Lee}, {and} \bibinfo{person}{Erik Learned-Miller}.}
  \bibinfo{year}{2013}\natexlab{}.
\newblock \showarticletitle{Augmenting CRFs with Boltzmann machine shape priors
  for image labeling}. In \bibinfo{booktitle}{\emph{Proceedings of the IEEE
  conference on computer vision and pattern recognition}}.
  \bibinfo{pages}{2019--2026}.
\newblock


\bibitem[\protect\citeauthoryear{Kemelmacher-Shlizerman, Seitz, Miller, and
  Brossard}{Kemelmacher-Shlizerman et~al\mbox{.}}{2016}]%
        {kemelmacher2016megaface}
\bibfield{author}{\bibinfo{person}{Ira Kemelmacher-Shlizerman},
  \bibinfo{person}{Steven~M Seitz}, \bibinfo{person}{Daniel Miller}, {and}
  \bibinfo{person}{Evan Brossard}.} \bibinfo{year}{2016}\natexlab{}.
\newblock \showarticletitle{The megaface benchmark: 1 million faces for
  recognition at scale}. In \bibinfo{booktitle}{\emph{Proceedings of the IEEE
  Conference on Computer Vision and Pattern Recognition}}.
  \bibinfo{pages}{4873--4882}.
\newblock


\bibitem[\protect\citeauthoryear{Koestinger, Wohlhart, Roth, and
  Bischof}{Koestinger et~al\mbox{.}}{2011}]%
        {koestinger2011annotated}
\bibfield{author}{\bibinfo{person}{Martin Koestinger}, \bibinfo{person}{Paul
  Wohlhart}, \bibinfo{person}{Peter~M Roth}, {and} \bibinfo{person}{Horst
  Bischof}.} \bibinfo{year}{2011}\natexlab{}.
\newblock \showarticletitle{Annotated facial landmarks in the wild: A
  large-scale, real-world database for facial landmark localization}. In
  \bibinfo{booktitle}{\emph{2011 IEEE international conference on computer
  vision workshops (ICCV workshops)}}. IEEE, \bibinfo{pages}{2144--2151}.
\newblock


\bibitem[\protect\citeauthoryear{Lai, Xiao, Pan, Cui, Feng, Xu, Yin, and
  Yan}{Lai et~al\mbox{.}}{2018}]%
        {lai2018deep}
\bibfield{author}{\bibinfo{person}{Hanjiang Lai}, \bibinfo{person}{Shengtao
  Xiao}, \bibinfo{person}{Yan Pan}, \bibinfo{person}{Zhen Cui},
  \bibinfo{person}{Jiashi Feng}, \bibinfo{person}{Chunyan Xu},
  \bibinfo{person}{Jian Yin}, {and} \bibinfo{person}{Shuicheng Yan}.}
  \bibinfo{year}{2018}\natexlab{}.
\newblock \showarticletitle{Deep recurrent regression for facial landmark
  detection}.
\newblock \bibinfo{journal}{\emph{IEEE Transactions on Circuits and Systems for
  Video Technology}} \bibinfo{volume}{28}, \bibinfo{number}{5}
  (\bibinfo{year}{2018}), \bibinfo{pages}{1144--1157}.
\newblock


\bibitem[\protect\citeauthoryear{Le, Brandt, Lin, Bourdev, and Huang}{Le
  et~al\mbox{.}}{2012}]%
        {le2012interactive}
\bibfield{author}{\bibinfo{person}{Vuong Le}, \bibinfo{person}{Jonathan
  Brandt}, \bibinfo{person}{Zhe Lin}, \bibinfo{person}{Lubomir Bourdev}, {and}
  \bibinfo{person}{Thomas~S Huang}.} \bibinfo{year}{2012}\natexlab{}.
\newblock \showarticletitle{Interactive facial feature localization}. In
  \bibinfo{booktitle}{\emph{European conference on computer vision}}. Springer,
  \bibinfo{pages}{679--692}.
\newblock


\bibitem[\protect\citeauthoryear{Levin, Rav-Acha, and Lischinski}{Levin
  et~al\mbox{.}}{2008}]%
        {levin2008spectral}
\bibfield{author}{\bibinfo{person}{Anat Levin}, \bibinfo{person}{Alex
  Rav-Acha}, {and} \bibinfo{person}{Dani Lischinski}.}
  \bibinfo{year}{2008}\natexlab{}.
\newblock \showarticletitle{Spectral matting}.
\newblock \bibinfo{journal}{\emph{IEEE transactions on pattern analysis and
  machine intelligence}} \bibinfo{volume}{30}, \bibinfo{number}{10}
  (\bibinfo{year}{2008}), \bibinfo{pages}{1699--1712}.
\newblock


\bibitem[\protect\citeauthoryear{Lin, Milan, Shen, and Reid}{Lin
  et~al\mbox{.}}{2017}]%
        {lin2017refinenet}
\bibfield{author}{\bibinfo{person}{Guosheng Lin}, \bibinfo{person}{Anton
  Milan}, \bibinfo{person}{Chunhua Shen}, {and} \bibinfo{person}{Ian Reid}.}
  \bibinfo{year}{2017}\natexlab{}.
\newblock \showarticletitle{Refinenet: Multi-path refinement networks for
  high-resolution semantic segmentation}. In
  \bibinfo{booktitle}{\emph{Proceedings of the IEEE conference on computer
  vision and pattern recognition}}. \bibinfo{pages}{1925--1934}.
\newblock


\bibitem[\protect\citeauthoryear{Liu, Shi, Liang, and Yang}{Liu
  et~al\mbox{.}}{2017}]%
        {liu2017face}
\bibfield{author}{\bibinfo{person}{Sifei Liu}, \bibinfo{person}{Jianping Shi},
  \bibinfo{person}{Ji Liang}, {and} \bibinfo{person}{Ming-Hsuan Yang}.}
  \bibinfo{year}{2017}\natexlab{}.
\newblock \showarticletitle{Face parsing via recurrent propagation}.
\newblock \bibinfo{journal}{\emph{arXiv preprint arXiv:1708.01936}}
  (\bibinfo{year}{2017}).
\newblock


\bibitem[\protect\citeauthoryear{Liu, Yang, Huang, and Yang}{Liu
  et~al\mbox{.}}{2015}]%
        {liu2015multi}
\bibfield{author}{\bibinfo{person}{Sifei Liu}, \bibinfo{person}{Jimei Yang},
  \bibinfo{person}{Chang Huang}, {and} \bibinfo{person}{Ming-Hsuan Yang}.}
  \bibinfo{year}{2015}\natexlab{}.
\newblock \showarticletitle{Multi-objective convolutional learning for face
  labeling}. In \bibinfo{booktitle}{\emph{Proceedings of the IEEE Conference on
  Computer Vision and Pattern Recognition}}. \bibinfo{pages}{3451--3459}.
\newblock


\bibitem[\protect\citeauthoryear{Long, Shelhamer, and Darrell}{Long
  et~al\mbox{.}}{2015}]%
        {long2015fully}
\bibfield{author}{\bibinfo{person}{Jonathan Long}, \bibinfo{person}{Evan
  Shelhamer}, {and} \bibinfo{person}{Trevor Darrell}.}
  \bibinfo{year}{2015}\natexlab{}.
\newblock \showarticletitle{Fully convolutional networks for semantic
  segmentation}. In \bibinfo{booktitle}{\emph{Proceedings of the IEEE
  conference on computer vision and pattern recognition}}.
  \bibinfo{pages}{3431--3440}.
\newblock


\bibitem[\protect\citeauthoryear{Luo, Wang, and Tang}{Luo
  et~al\mbox{.}}{2012}]%
        {luo2012hierarchical}
\bibfield{author}{\bibinfo{person}{Ping Luo}, \bibinfo{person}{Xiaogang Wang},
  {and} \bibinfo{person}{Xiaoou Tang}.} \bibinfo{year}{2012}\natexlab{}.
\newblock \showarticletitle{Hierarchical face parsing via deep learning}. In
  \bibinfo{booktitle}{\emph{2012 IEEE Conference on Computer Vision and Pattern
  Recognition}}. IEEE, \bibinfo{pages}{2480--2487}.
\newblock


\bibitem[\protect\citeauthoryear{Merget, Rock, and Rigoll}{Merget
  et~al\mbox{.}}{2018}]%
        {merget2018robust}
\bibfield{author}{\bibinfo{person}{Daniel Merget}, \bibinfo{person}{Matthias
  Rock}, {and} \bibinfo{person}{Gerhard Rigoll}.}
  \bibinfo{year}{2018}\natexlab{}.
\newblock \showarticletitle{Robust facial landmark detection via a
  fully-convolutional local-global context network}. In
  \bibinfo{booktitle}{\emph{Proceedings of the IEEE Conference on Computer
  Vision and Pattern Recognition}}. \bibinfo{pages}{781--790}.
\newblock


\bibitem[\protect\citeauthoryear{Messer, Matas, Kittler, Luettin, and
  Maitre}{Messer et~al\mbox{.}}{1999}]%
        {messer1999xm2vtsdb}
\bibfield{author}{\bibinfo{person}{Kieron Messer}, \bibinfo{person}{Jiri
  Matas}, \bibinfo{person}{Josef Kittler}, \bibinfo{person}{Juergen Luettin},
  {and} \bibinfo{person}{Gilbert Maitre}.} \bibinfo{year}{1999}\natexlab{}.
\newblock \showarticletitle{XM2VTSDB: The extended M2VTS database}. In
  \bibinfo{booktitle}{\emph{Second international conference on audio and
  video-based biometric person authentication}}, Vol.~\bibinfo{volume}{964}.
  \bibinfo{pages}{965--966}.
\newblock


\bibitem[\protect\citeauthoryear{Newell, Yang, and Deng}{Newell
  et~al\mbox{.}}{2016}]%
        {newell2016stacked}
\bibfield{author}{\bibinfo{person}{Alejandro Newell}, \bibinfo{person}{Kaiyu
  Yang}, {and} \bibinfo{person}{Jia Deng}.} \bibinfo{year}{2016}\natexlab{}.
\newblock \showarticletitle{Stacked hourglass networks for human pose
  estimation}. In \bibinfo{booktitle}{\emph{European Conference on Computer
  Vision}}. Springer, \bibinfo{pages}{483--499}.
\newblock


\bibitem[\protect\citeauthoryear{Ou, Liu, Cao, and Ling}{Ou
  et~al\mbox{.}}{2016}]%
        {ou2016beauty}
\bibfield{author}{\bibinfo{person}{Xinyu Ou}, \bibinfo{person}{Si Liu},
  \bibinfo{person}{Xiaochun Cao}, {and} \bibinfo{person}{Hefei Ling}.}
  \bibinfo{year}{2016}\natexlab{}.
\newblock \showarticletitle{Beauty emakeup: A deep makeup transfer system}. In
  \bibinfo{booktitle}{\emph{Proceedings of the 24th ACM international
  conference on Multimedia}}. ACM, \bibinfo{pages}{701--702}.
\newblock


\bibitem[\protect\citeauthoryear{Ramanan and Zhu}{Ramanan and Zhu}{2012}]%
        {ramanan2012face}
\bibfield{author}{\bibinfo{person}{Deva Ramanan} {and}
  \bibinfo{person}{Xiangxin Zhu}.} \bibinfo{year}{2012}\natexlab{}.
\newblock \showarticletitle{Face detection, pose estimation, and landmark
  localization in the wild}. In \bibinfo{booktitle}{\emph{2012 IEEE conference
  on computer vision and pattern recognition}}. IEEE,
  \bibinfo{pages}{2879--2886}.
\newblock


\bibitem[\protect\citeauthoryear{Sagonas, Tzimiropoulos, Zafeiriou, and
  Pantic}{Sagonas et~al\mbox{.}}{2013}]%
        {sagonas2013300}
\bibfield{author}{\bibinfo{person}{Christos Sagonas}, \bibinfo{person}{Georgios
  Tzimiropoulos}, \bibinfo{person}{Stefanos Zafeiriou}, {and}
  \bibinfo{person}{Maja Pantic}.} \bibinfo{year}{2013}\natexlab{}.
\newblock \showarticletitle{300 faces in-the-wild challenge: The first facial
  landmark localization challenge}. In \bibinfo{booktitle}{\emph{Proceedings of
  the IEEE International Conference on Computer Vision Workshops}}.
  \bibinfo{pages}{397--403}.
\newblock


\bibitem[\protect\citeauthoryear{Smith, Zhang, Brandt, Lin, and Yang}{Smith
  et~al\mbox{.}}{2013}]%
        {smith2013exemplar}
\bibfield{author}{\bibinfo{person}{Brandon~M Smith}, \bibinfo{person}{Li
  Zhang}, \bibinfo{person}{Jonathan Brandt}, \bibinfo{person}{Zhe Lin}, {and}
  \bibinfo{person}{Jianchao Yang}.} \bibinfo{year}{2013}\natexlab{}.
\newblock \showarticletitle{Exemplar-based face parsing}. In
  \bibinfo{booktitle}{\emph{Proceedings of the IEEE Conference on Computer
  Vision and Pattern Recognition}}. \bibinfo{pages}{3484--3491}.
\newblock


\bibitem[\protect\citeauthoryear{Sun, Wang, and Tang}{Sun
  et~al\mbox{.}}{2013}]%
        {sun2013deep}
\bibfield{author}{\bibinfo{person}{Yi Sun}, \bibinfo{person}{Xiaogang Wang},
  {and} \bibinfo{person}{Xiaoou Tang}.} \bibinfo{year}{2013}\natexlab{}.
\newblock \showarticletitle{Deep convolutional network cascade for facial point
  detection}. In \bibinfo{booktitle}{\emph{Proceedings of the IEEE conference
  on computer vision and pattern recognition}}. \bibinfo{pages}{3476--3483}.
\newblock


\bibitem[\protect\citeauthoryear{Tao~Ruan}{Tao~Ruan}{2018}]%
        {CE2P2019}
\bibfield{author}{\bibinfo{person}{Zilong Huang Yunchao Wei Shikui Wei Yao Zhao
  Thomas~Huang Tao~Ruan, Ting~Liu}.} \bibinfo{year}{2018}\natexlab{}.
\newblock \showarticletitle{Devil in the Details: Towards Accurate Single and
  Multiple Human Parsing}.
\newblock \bibinfo{journal}{\emph{arXiv:1809.05996}} (\bibinfo{year}{2018}).
\newblock


\bibitem[\protect\citeauthoryear{Warrell and Prince}{Warrell and
  Prince}{2009}]%
        {warrell2009labelfaces}
\bibfield{author}{\bibinfo{person}{Jonathan Warrell} {and}
  \bibinfo{person}{Simon~JD Prince}.} \bibinfo{year}{2009}\natexlab{}.
\newblock \showarticletitle{Labelfaces: Parsing facial features by multiclass
  labeling with an epitome prior}. In \bibinfo{booktitle}{\emph{2009 16th IEEE
  international conference on image processing (ICIP)}}. IEEE,
  \bibinfo{pages}{2481--2484}.
\newblock


\bibitem[\protect\citeauthoryear{Wei, Sun, Wang, Lai, and Liu}{Wei
  et~al\mbox{.}}{2017}]%
        {wei2017learning}
\bibfield{author}{\bibinfo{person}{Zhen Wei}, \bibinfo{person}{Yao Sun},
  \bibinfo{person}{Jinqiao Wang}, \bibinfo{person}{Hanjiang Lai}, {and}
  \bibinfo{person}{Si Liu}.} \bibinfo{year}{2017}\natexlab{}.
\newblock \showarticletitle{Learning adaptive receptive fields for deep image
  parsing network}. In \bibinfo{booktitle}{\emph{Proceedings of the IEEE
  Conference on Computer Vision and Pattern Recognition}}.
  \bibinfo{pages}{2434--2442}.
\newblock


\bibitem[\protect\citeauthoryear{Wu, Qian, Yang, Wang, Cai, and Zhou}{Wu
  et~al\mbox{.}}{2018}]%
        {wu2018look}
\bibfield{author}{\bibinfo{person}{Wayne Wu}, \bibinfo{person}{Chen Qian},
  \bibinfo{person}{Shuo Yang}, \bibinfo{person}{Quan Wang},
  \bibinfo{person}{Yici Cai}, {and} \bibinfo{person}{Qiang Zhou}.}
  \bibinfo{year}{2018}\natexlab{}.
\newblock \showarticletitle{Look at boundary: A boundary-aware face alignment
  algorithm}. In \bibinfo{booktitle}{\emph{Proceedings of the IEEE Conference
  on Computer Vision and Pattern Recognition}}. \bibinfo{pages}{2129--2138}.
\newblock


\bibitem[\protect\citeauthoryear{Xiong and De~la Torre}{Xiong and De~la
  Torre}{2013}]%
        {xiong2013supervised}
\bibfield{author}{\bibinfo{person}{Xuehan Xiong} {and}
  \bibinfo{person}{Fernando De~la Torre}.} \bibinfo{year}{2013}\natexlab{}.
\newblock \showarticletitle{Supervised descent method and its applications to
  face alignment}. In \bibinfo{booktitle}{\emph{Proceedings of the IEEE
  conference on computer vision and pattern recognition}}.
  \bibinfo{pages}{532--539}.
\newblock


\bibitem[\protect\citeauthoryear{Yamashita, Nakamura, Fukui, Yamauchi, and
  Fujiyoshi}{Yamashita et~al\mbox{.}}{2015}]%
        {yamashita2015cost}
\bibfield{author}{\bibinfo{person}{Takayoshi Yamashita},
  \bibinfo{person}{Takaya Nakamura}, \bibinfo{person}{Hiroshi Fukui},
  \bibinfo{person}{Yuji Yamauchi}, {and} \bibinfo{person}{Hironobu Fujiyoshi}.}
  \bibinfo{year}{2015}\natexlab{}.
\newblock \showarticletitle{Cost-alleviative learning for deep convolutional
  neural network-based facial part labeling}.
\newblock \bibinfo{journal}{\emph{IPSJ Transactions on Computer Vision and
  Applications}}  \bibinfo{volume}{7} (\bibinfo{year}{2015}),
  \bibinfo{pages}{99--103}.
\newblock


\bibitem[\protect\citeauthoryear{Yang, Liu, and Zhang}{Yang
  et~al\mbox{.}}{2017}]%
        {yang2017stacked}
\bibfield{author}{\bibinfo{person}{Jing Yang}, \bibinfo{person}{Qingshan Liu},
  {and} \bibinfo{person}{Kaihua Zhang}.} \bibinfo{year}{2017}\natexlab{}.
\newblock \showarticletitle{Stacked hourglass network for robust facial
  landmark localisation}. In \bibinfo{booktitle}{\emph{Proceedings of the IEEE
  Conference on Computer Vision and Pattern Recognition Workshops}}.
  \bibinfo{pages}{79--87}.
\newblock


\bibitem[\protect\citeauthoryear{Zhang, Riggan, Hu, Short, and Patel}{Zhang
  et~al\mbox{.}}{2018}]%
        {zhang2018synthesis}
\bibfield{author}{\bibinfo{person}{He Zhang}, \bibinfo{person}{Benjamin~S
  Riggan}, \bibinfo{person}{Shuowen Hu}, \bibinfo{person}{Nathaniel~J Short},
  {and} \bibinfo{person}{Vishal~M Patel}.} \bibinfo{year}{2018}\natexlab{}.
\newblock \showarticletitle{Synthesis of High-Quality Visible Faces from
  Polarimetric Thermal Faces using Generative Adversarial Networks}.
\newblock \bibinfo{journal}{\emph{International Journal of Computer Vision}}
  (\bibinfo{year}{2018}), \bibinfo{pages}{1--18}.
\newblock


\bibitem[\protect\citeauthoryear{Zhang, Luo, Loy, and Tang}{Zhang
  et~al\mbox{.}}{2014}]%
        {zhang2014facial}
\bibfield{author}{\bibinfo{person}{Zhanpeng Zhang}, \bibinfo{person}{Ping Luo},
  \bibinfo{person}{Chen~Change Loy}, {and} \bibinfo{person}{Xiaoou Tang}.}
  \bibinfo{year}{2014}\natexlab{}.
\newblock \showarticletitle{Facial landmark detection by deep multi-task
  learning}. In \bibinfo{booktitle}{\emph{European conference on computer
  vision}}. Springer, \bibinfo{pages}{94--108}.
\newblock


\bibitem[\protect\citeauthoryear{Zhao, Shi, Qi, Wang, and Jia}{Zhao
  et~al\mbox{.}}{2017}]%
        {zhao2017pyramid}
\bibfield{author}{\bibinfo{person}{Hengshuang Zhao}, \bibinfo{person}{Jianping
  Shi}, \bibinfo{person}{Xiaojuan Qi}, \bibinfo{person}{Xiaogang Wang}, {and}
  \bibinfo{person}{Jiaya Jia}.} \bibinfo{year}{2017}\natexlab{}.
\newblock \showarticletitle{Pyramid scene parsing network}. In
  \bibinfo{booktitle}{\emph{Proceedings of the IEEE conference on computer
  vision and pattern recognition}}. \bibinfo{pages}{2881--2890}.
\newblock


\bibitem[\protect\citeauthoryear{Zhou, Fan, Cao, Jiang, and Yin}{Zhou
  et~al\mbox{.}}{2013}]%
        {zhou2013extensive}
\bibfield{author}{\bibinfo{person}{Erjin Zhou}, \bibinfo{person}{Haoqiang Fan},
  \bibinfo{person}{Zhimin Cao}, \bibinfo{person}{Yuning Jiang}, {and}
  \bibinfo{person}{Qi Yin}.} \bibinfo{year}{2013}\natexlab{}.
\newblock \showarticletitle{Extensive facial landmark localization with
  coarse-to-fine convolutional network cascade}. In
  \bibinfo{booktitle}{\emph{Proceedings of the IEEE International Conference on
  Computer Vision Workshops}}. \bibinfo{pages}{386--391}.
\newblock


\bibitem[\protect\citeauthoryear{Zhu, Lei, Liu, Shi, and Li}{Zhu
  et~al\mbox{.}}{2016}]%
        {zhu2016face}
\bibfield{author}{\bibinfo{person}{Xiangyu Zhu}, \bibinfo{person}{Zhen Lei},
  \bibinfo{person}{Xiaoming Liu}, \bibinfo{person}{Hailin Shi}, {and}
  \bibinfo{person}{Stan~Z Li}.} \bibinfo{year}{2016}\natexlab{}.
\newblock \showarticletitle{Face alignment across large poses: A 3d solution}.
  In \bibinfo{booktitle}{\emph{Proceedings of the IEEE conference on computer
  vision and pattern recognition}}. \bibinfo{pages}{146--155}.
\newblock


\end{thebibliography}

\end{document}